\documentclass[sigconf,acm]{acmart}

\usepackage{mdframed}
\usepackage{tikz}
\usepackage{todonotes}
\newmdtheoremenv[
  linewidth=0.5pt,
  linecolor=black!50,
  leftmargin=0pt,
  rightmargin=0pt,
  innertopmargin=8pt,      
  innerbottommargin=8pt,   
  innerleftmargin=10pt,    
  innerrightmargin=10pt,   
  skipabove=10pt,          
  skipbelow=12pt,          
]{definition}{Definition}
\AtBeginDocument{%
  }

\NewDocumentCommand{\cheng} { mO{} }{\textcolor{purple}{\textsuperscript{\textit{Cheng}}\textsf{\textbf{\small[#1]}}}}

\copyrightyear{2026}
\acmYear{2026}
\setcopyright{cc}
\setcctype{by}
\acmConference[AI Summit '26]{ACM AI Summit 2026}{August 31-September 02, 2026}{Atlanta, GA, USA}
\acmBooktitle{ACM AI Summit 2026 (AI Summit '26), August 31-September 02, 2026, Atlanta, GA, USA}
\acmDOI{10.1145/3806096.3844850}
\acmISBN{979-8-4007-2638-5/2026/08}

\begin{document}

\title{Generative Interpretability via Scalable Neuro-Symbolic Models}

\author{Xiaocong Yang}
\affiliation{\institution{AI Interpretability @ Illinois, University of Illinois Urbana-Champaign}\city{Urbana}\state{IL}\country{USA}}
\email{xy51@illinois.edu}


\begin{abstract}
As the use of Large Language Models moves from chatbots into agentic systems, where outputs become actions with irreversible consequences on reality, the existing paradigm on AI Interpretability research, post-hoc interpretability, is structurally inadequate for safe and trustworthy model deployment: it explains behavior after the fact but cannot audit or intervene in an inference computation before it commits to an output. We therefore argue for a shift toward \emph{generative interpretability}, an architectural property under which a model's inference pass natively exposes semantically meaningful checkpoints that are human-understandable and amenable to causal intervention. We show the merits of generative interpretability as comparison to other interpretability research paradigms, and propose Neuro-Symbolic Models as a concrete instantiation. 
\end{abstract}



\maketitle

\section{Introduction}

Recent years have witnessed the rapid development of connectionist AI models, especially Large Language Models (LLMs), with unprecedented real-world impacts. Although powerful, they are generally believed to lack interpretability of their internal mechanisms and are therefore described as "black-box" models \cite{lipton2017mythosmodelinterpretability, rudin2019stopexplainingblackbox}. As a direct consequence, it is usually infeasible to understand, audit and intervene in undesirable behaviors such as hallucination, sycophancy and harmful content generation in a scalable and systematic way. These consequences are dramatically amplified as LLMs are deployed in agentic systems, such as healthcare systems that help with diagnoses \cite{wang2025safetychallengesaimedicine} and autonomous driving systems that control a running vehicle \cite{wu2025vulnerabilityllmvlmcontrolledrobotics}. There, agents interact with the real world and can cause irreversible consequences far more severe than in a chatbot: as reported by \citet{stanfordAIindex2025}, documented AI safety incidents surged 56.4\% from 2023 to 2024, with autonomous driving systems involved in fatal crashes~\cite{aiincidents2024} and agentic AI systems causing financial losses~\cite{internationalAIsafety2026}. The issue has risen to the level of national security: the DARPA-NSF-CAISI AI Forge program~\cite{darpaAIForge2026} identifies AI Interpretability as one of three priorities in AI research, highlighting interpretable AI in agentic settings for auditable autonomy and operational runtime control.


In this work, we argue that a paradigm transition is necessary for interpretability research to meet this requirement. Most current work on explaining LLMs falls into post-hoc interpretability \cite{Madsen_2022}, which relies on reverse-engineering tools external to the LLM. Such tools are confined to observation: their interpretations are confounded by the tools' own approximation errors, and the tools are usually discarded at deployment rather than used to make the deployed model safer. Post-hoc interpretability thus cannot reliably translate into safer behaviors of deployed LLMs. Instead, we propose \textit{generative interpretability} as an alternative, with the key claim that \textbf{interpretability should be a property inherent to the model computations that allows effective audit and intervention at deployment,} directly targeting the research thrusts and challenges in the DARPA-NSF-CAISI AI Forge report. 


\section{Post-hoc Interpretability and Its Limitations}
\label{section2}
Post-hoc interpretability attributes and grounds the computations and behaviors of an already-trained model into human understandable terms \cite{Madsen_2022}. By explanandum, it falls into two categories: \textit{anatomical interpretability} and \textit{physiological interpretability}.

\subsection{Anatomical Interpretability}
Anatomical interpretability studies \textit{local structures} in a model that function stably and consistently \textit{over different inputs}, usually accompanied by modularity and functional specialization. For example, the multi-head self-attention layers (MHAs) and feed-forward networks (FFNs) in a Transformer \cite{vaswani2023attentionneed} were thoroughly studied before the emergence of LLMs: the MHAs in a BERT model \cite{bert} specialize in processing certain semantic or syntactic information \cite{clark-etal-2019-bert}, and the FFNs are interpreted as factual knowledge banks \cite{geva-etal-2021-transformer, dai-etal-2022-knowledge} that can be edited to change model outputs on certain facts \cite{meng2023locatingeditingfactualassociations}. 

\subsection{Physiological Interpretability}
Physiological interpretability recovers the \textit{end-to-end information flow} of a \textit{specific inference pass} executed by the model. Its theoretical basis traces back to causal mediation analysis \cite{pearl2013directindirecteffects}, where the computation on a given input is characterized as a Directed Acyclic Graph (DAG) and interventions on intermediate nodes reveal their causal effects on model output. Early efforts include probing \cite{belinkov2021probingclassifierspromisesshortcomings} and naive saliency tracing \cite{sundararajan2017axiomaticattributiondeepnetworks, Li_2019, sundararajan2020shapleyvaluesmodelexplanation}. With the development of LLMs, the Transformer Circuit Theory \cite{elhage2021mathematical} and its follow-up work \cite{elhage2022superposition, bricken2023monosemanticity, lindsey2025biology} have become among the most cutting-edge post-hoc techniques. Instead of naive, metric-based tracing, they reduce an inference pass of an LLM to combinations of simplified operations on semantically meaningful directions in activation spaces, called \textit{features}, and answer three key questions: (1) how features are distributed in activation spaces (quasi-orthogonally); (2) how to identify features and their meanings (via dictionary learning); and (3) how a model uses these features to generate its output (via circuit tracing with attribution graphs). 

\subsection{The Shared Fundamental Limitation}

Post-hoc methods give us insights into LLMs, but all of them share a fundamental limitation: \textbf{they can explain what a model does, but cannot change what it does, in a systematic and scalable way.} On the one hand, post-hoc tools are far less effective in inference-time intervention than in explaining representations. For example, \citet{basu2026interpretabilityactionabilitymechanisticmethods} documented a large gap between a model's internal representations and its outputs that steering interventions based on various post-hoc tools failed to reliably close. On the other hand, post-hoc tools suffer from scalability and/or efficiency problems that block their deployment outside the laboratory. Powerful tools in Transformer Circuit Theory, such as Sparse Autoencoders and Cross-layer Transcoders, cannot be efficiently integrated into deployed LLMs due to the extremely large feature dimensions needed for sparse and clean decomposition. Lightweight tools, such as model editing \cite{wang2024knowledgeeditinglargelanguage} and activation patching \cite{zhang2024bestpracticesactivationpatching}, are limited to modifying a specific type of behavior and depend on labeled data for it, and are hence far from scalable and systematic. 
\section{Generative Interpretability}
\label{section3}

The failure of post-hoc tools at intervention motivates a new interpretability paradigm tailored for effective audit. In this section, we articulate generative interpretability as the requirement on models that enables inference-time audit and controllable intervention. 

\subsection{Definition}
We define generative interpretability as follows.

\begin{definition}[Generative Interpretability]
\label{def:gi}
An architectural property of a model under which its inference pass\footnotemark{} exposes intermediate checkpoints that are (i) semantically meaningful to humans natively and (ii) causally intervenable by construction.
\end{definition}
\footnotetext{We use the term ``inference pass'' to describe a single computation trajectory that maps input to a single meaningful output unit. For example, in auto-regressive LLMs, it is the generation of a next token. We use ``inference path" for the static computational pathway defined by the model architecture, along which an inference pass is executed for a given input. } 
\vspace{6pt}

\textbf{Human understandability. } Interpretability in AI is typically framed in terms of human social attribution and cognition \cite{Biran2017ExplanationAJ, miller}. By \emph{semantically meaningful to humans}, we require a checkpoint to map to a human-readable concept. Audit and intervention ultimately reflect human decisions and values, and can only identify undesired behaviors when intermediate states are expressed in terms humans understand. This excludes interpretations that are only transparent as mathematical functions. For instance, an optimization view yields a mathematically interpretable Transformer \cite{yu2023whiteboxtransformerssparserate} whose layers have well-defined roles such as token set compression and sparsification, yet it reveals no semantic meaning the model assigns to its inputs. By \emph{native}, we require that meanings are acquired without external recovery tools. For example, the framework by \citet{mueller2025questrightmediatorsurveying} separates the choice of checkpoints from the discovery of their meanings; whether the checkpoints are Attention heads, neurons or activation subspaces, their meanings are searched and assigned by post-hoc tools, inheriting the key defects of post-hoc interpretability. 

\textbf{Causal intervention.} An interpretability framework is considered faithful when it reflects the real \emph{reasoning process} of a model \cite{jacovi2020faithfullyinterpretablenlpsystems, geiger2025causalabstractiontheoreticalfoundation}. In particular, it should be testable with \emph{``do-intervention''} \cite{causalinference}, with counterfactual outputs consistently aligned with the interpretation, as argued in \cite{nussbaumhoffer2026llmexplainabilitycounterfactualchains}. This is essential for predictable model behaviors under intervention, which post-hoc tools usually fail to provide. It implies that checkpoints must be variables the inference pass functionally depends on: altering their values necessarily propagates through downstream computation, so do-interventions have well-defined counterfactuals by construction, rather than the contingent and approximate causal effects of post-hoc tools.

\subsection{Two Axes of Interpretability}
To relate generative interpretability to existing paradigms, we compare them along two axes: \emph{when} interpretation is available relative to inference (the temporal axis), and \emph{where} along the inference path human-readable structures reside (the spatial axis).

\textbf{The temporal axis.} We divide interpretability approaches into three categories by when human-understandable states are available relative to inference. Most traditional machine learning models, such as linear regression, decision trees, and rule-based systems, are interpretable \textit{before inference}: their structure or static parameters are human-understandable and fully determine their behavior on every input, so inference yields no interpretable information beyond applying already-readable patterns. \citet{rudin2019stopexplainingblackbox} advocates this property, and follow-up work shows the superior performance of a fully interpretable, domain-specific model \cite{hu2023optimalsparsedecisiontrees}. Such models are usually called \textit{white-box models}. Post-hoc methods recover interpretable states only \textit{after inference}: the states are not natively exposed by the model but reconstructed by external tools from completed inference passes. The third category, where interpretable states emerge during inference, has drawn less attention. Representatives are Concept Bottleneck Models (CBMs) \cite{koh2020conceptbottleneckmodels} and their extension to LLMs \cite{sun2025conceptbottlenecklargelanguage}, which ground neural computations in a pre-defined, human-understandable concept pool as the final component of the inference path, but are usually limited to simple similarity matching in a specific domain. 

\textbf{The spatial axis.} Approaches can also be categorized by the spatial position of interpretable states in an inference path. As a baseline, vanilla LLMs and interpretations based on them (such as Chain-of-thought \cite{wei2023chainofthoughtpromptingelicitsreasoning}) offer human-readable text tokens only \textit{at input/output boundaries}. Post-hoc methods construct interpretations with tools that lie \textit{externally} to the model. The CBM family provides an interpretable \textit{interior checkpoint layer} for simple concept matching, and most white-box models are interpretable \textit{throughout the model} by definition.

\textbf{The position of generative interpretability.} Along both axes, generative interpretability occupies an intermediate position. Temporally, it resides \textbf{during inference}: a checkpoint's schema (concept space) is usually pre-defined, while the values filling it are produced by an opaque sub-computation (such as a neural network) as the input is processed. The interpretable states thus can neither be derived from the static model alone nor require the full inference pass to complete. Spatially, it requires \textbf{multiple interior checkpoints} for effective causal intervention: boundary readability exposes only inputs and outputs, and a single checkpoint captures only one cross-section of a multi-step computation, which is insufficient to attribute undesired outputs (Section \ref{section3.3}). 

\subsection{Generative Interpretability as the Minimal Sufficient Commitment for Auditable LLMs}
\label{section3.3}

We argue that, for LLMs, generative interpretability is the minimal architectural commitment sufficient for audit and intervention over the inference pass that is scalable in development and tractable at deployment. The commitment is prescriptive: a property of the models we should build upon existing LLMs, not of LLMs as they are. 

\textbf{A partial order by auditable content.} The axes induce partial orders on commitment strength. Spatially, a throughout-interpretable model is legible at any interior point, subsuming multiple-checkpoint readability, which subsumes a single checkpoint and then the input-output boundary. Temporally, a model legible before inference remains legible during it when given an input. Post-hoc interpretability lies outside this chain: its readable structure is reconstructed externally rather than residing in the model, implying no architectural commitment.

\textbf{Why not weaker.} Auditing content before it reaches the output immediately rules out after-inference interpretability, whose readable state is recovered only once the output—and any unsafe content—already exists. During inference is thus the weakest temporal position that still permits pre-generation audit. Spatially, two weaker commitments fall short. The input-output boundary (vanilla LLMs, CoT) exposes no intermediate checkpoints on the inference path. Any audit of CoT's intermediate steps operates on tokens already generated by inference passes, so it can only reject an undesired generation rather than attribute it to the responsible computation and correct it. Moreover, CoT steps do not faithfully reflect the internal computation \cite{arcuschin2026chainofthoughtreasoningwildfaithful, scalena2026commitmentboundaryprobingepiphenomenal, turpin2023languagemodelsdontsay}, making intervention on them unreliable \cite{valmeekam2026beyond}. A single interior checkpoint as in CBMs audits one cross-section and, for LLMs, cannot attribute an undesired output to the computation either, as empirically verified in \cite{basu2026interpretabilityactionabilitymechanisticmethods}. Theoretically, \citet{poggio2025efficientlycomputablefunctionsdeep} show that any efficiently computable function, characterized by efficient Turing computability, is provably compositionally sparse with a DAG structure. Causal intervention as in Definition \ref{def:gi} therefore requires at least multiple checkpoints to model such structures.

\textbf{Why not stronger. } One might demand a stronger, white-box-style commitment. However, such semantic throughout-readability harms the representational capacity of the model. A representative effort is to achieve monosemanticity of neurons or directions in activation spaces for a clean mapping between semantic concepts and model structures. This is infeasible for LLMs, which must represent far more meaningful concepts than they have dimensions, packing them into near-orthogonal directions through superposition \cite{elhage2022superposition}. A $d$-dimensional space admits exponentially many such near-orthogonal directions \cite{Johnson1984ExtensionsOL}, and forcing every direction to be monosemantic would abandon this packing, collapsing the number of meaningful concepts from exponential to linear in the dimension. Generative interpretability, however, commits to interpretability only at limited checkpoints, leaving most of the model unconstrained to represent knowledge efficiently.

\section{Neuro-Symbolic Model as an Instance with Generative Interpretability}
\label{section4}

\begin{figure*}[t]
\centering
\includegraphics[width=0.85\textwidth, trim={8bp 8.5bp 16.5bp 111bp}, clip]{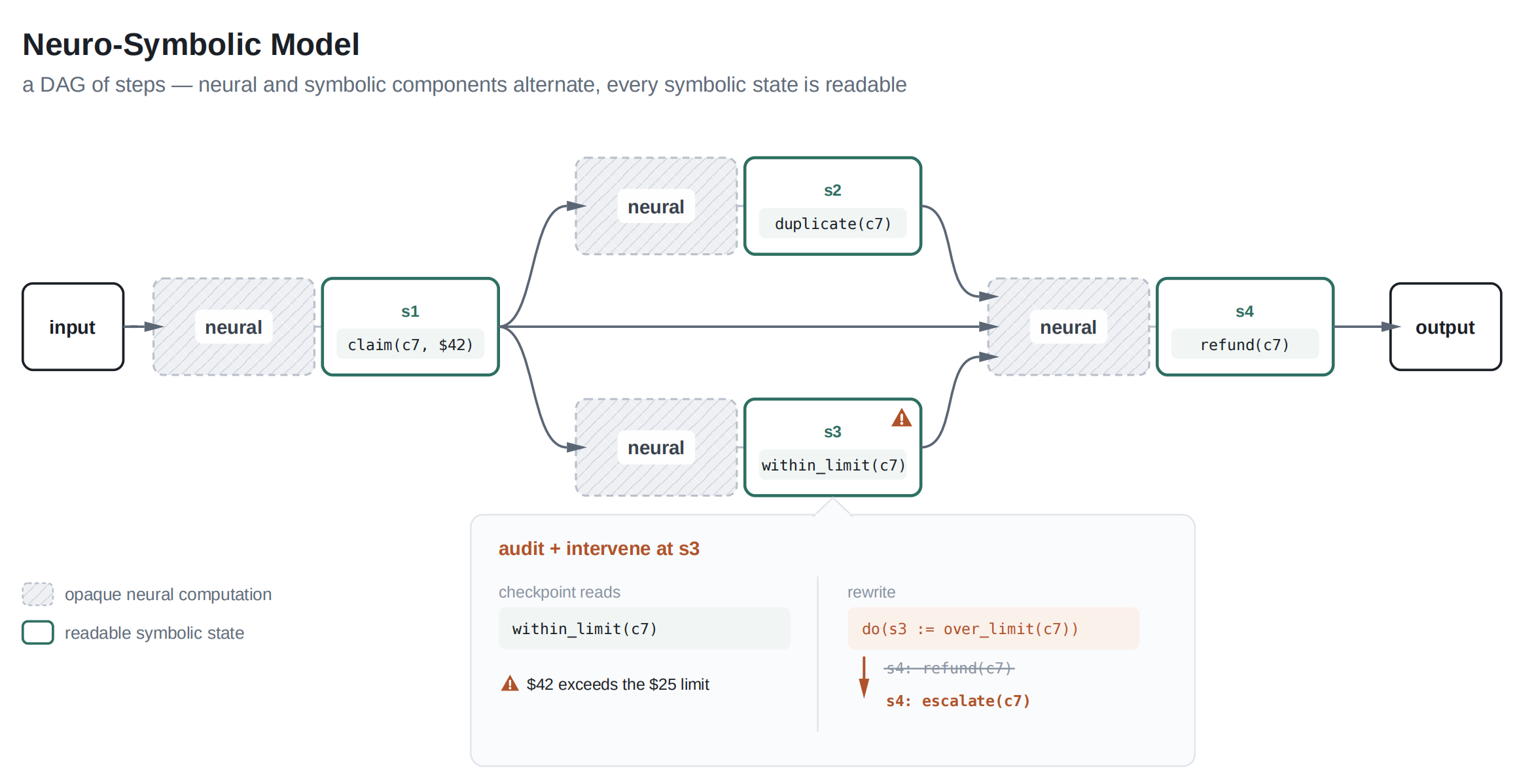}
\caption{An example of a Neuro-Symbolic Model handling a customer's refund request. Audit and intervene are designed at the refund amount checkpoint, as it is high-stake for a company.}
\Description{An inference DAG from input to output in which neural steps alternate with symbolic states: s1 claim(c7, \$42), s2 duplicate(c7), s3 within\_limit(c7), and s4 refund(c7). An audit panel at s3 shows that \$42 exceeds the \$25 limit; the auditor rewrites s3 to over\_limit(c7), which changes s4 from refund(c7) to escalate(c7).}
\label{fig:ns-refund}
\end{figure*}

In the history of AI, Neuro-Symbolic Models \cite{bhuyan2024neurosymbolic} are a successful instance that exploits the benefits of both the Connectionism and Symbolism paradigms. From the perspective of information compression, neural and symbolic models process raw input in fundamentally different ways. Symbolic systems compress by projecting onto a pre-defined schema vocabulary; their representations are discrete and hard-bounded, generalize algebraically \cite{marcus2001algebraic}, and naturally provide human-understandable states. Neural networks instead compress by projecting onto patterns learned during training, as continuous, distributed representations that generalize by interpolation. 

\begin{figure}[h]
\centering
\resizebox{\linewidth}{!}{%
\begin{tikzpicture}[
cell/.style={
    draw,
    minimum width=2cm,     
    minimum height=0.9cm,  
    align=center,
    font=\scriptsize,
    text width=1.8cm, 
    inner sep=2pt
},
  header/.style={
    font=\scriptsize\bfseries,
    align=center,
    text width=1.75cm
  },
  auditable/.style={
    cell,
    fill=black!25
  }
]
\node[header] at (1.95, 4.4) {Before Inference};
\node[header] at (3.95, 4.4) {During Inference};
\node[header] at (5.95, 4.4) {After Inference};
\node[header, anchor=east] at (0.85, 3.775) {Throughout};
\node[auditable] at (1.95, 3.775) {White-box Models \cite{rudin2019stopexplainingblackbox}};
\node[auditable] at (3.95, 3.775) {---};
\node[cell] at (5.95, 3.775) {---};
\node[header, anchor=east] at (0.85, 2.875) {Multiple Interior Checkpoints};
\node[auditable] at (1.95, 2.875) {---};
\node[auditable] at (3.95, 2.875) {\textbf{Neuro-Symbolic Models}};
\node[cell] at (5.95, 2.875) {---};
\node[header, anchor=east] at (0.85, 1.975) {Single Interior Checkpoint};
\node[cell] at (1.95, 1.975) {---};
\node[cell] at (3.95, 1.975) {CBM \cite{koh2020conceptbottleneckmodels}, CB-LLM \cite{sun2025conceptbottlenecklargelanguage}};
\node[cell] at (5.95, 1.975) {---};
\node[header, anchor=east] at (0.85, 1.075) {I/O Boundary Only};
\node[cell] at (1.95, 1.075) {---};
\node[cell] at (3.95, 1.075) {Vanilla LLM (CoT) \cite{wei2023chainofthoughtpromptingelicitsreasoning}};
\node[cell] at (5.95, 1.075) {---};
\node[header, anchor=east] at (0.85, 0.0) {External};
\node[cell, minimum height=1.25cm] at (1.95, 0.0) {---};
\node[cell, minimum height=1.25cm] at (3.95, 0.0) {---};
\node[cell, minimum height=1.25cm] at (5.95, 0.0) {Post-hoc Interp.\ (SAEs \cite{templeton2024scaling}, circuits \cite{lindsey2025biology}, probing \cite{belinkov2021probingclassifierspromisesshortcomings}, etc.)};
\draw[<-, thick] (1.0, 4.85) -- (6.9, 4.85);
\node[above, font=\scriptsize\itshape] at (3.95, 4.85) {Temporal (earlier)};
\draw[<-, thick] (-1.7, 4.2) -- (-1.7, -0.6);
\node[rotate=90, font=\scriptsize\itshape] at (-2.1, 1.8) {Spatial (more pervasive)};
\end{tikzpicture}%
}
\caption{Interpretability paradigms organized by two temporal and spatial axes. Shaded cells admit audit and intervention before undesired content is generated. Neuro-symbolic models realize generative interpretability within this region while preserving distributed, high-capacity representations.}

\label{fig:two-axes}
\end{figure}

\subsection{Symbolic Components as Checkpoints}
Unlike the vague patterns in neural networks, the representations in a symbolic system are human-understandable by construction. The causality requirement then selects the neuro-symbolic architectures that meet Definition \ref{def:gi}. Under the categorization of \cite{bhuyan2024neurosymbolic}, Type-3 Neuro-Symbolic Models, where neural and symbolic components cooperate in a feedback loop, best satisfy this requirement. A representative Type-3 model is NLProlog \cite{weber2019nlprologreasoningweakunification}, whose inference pass consists of composable proof steps represented by predicate symbols, while the similarity scoring by text embeddings inside each step stays opaque. The proof steps thus form multiple interior checkpoints within a single inference pass, each auditable against the expected intermediate representation. Figure~\ref{fig:ns-refund} illustrates such checkpoints in an agentic customer-service setting.

\subsection{Challenges of Building Neuro-Symbolic Models with LLMs}

The Neuro-Symbolic paradigm is a promising candidate to mitigate the black-box nature of LLMs through its intrinsic generative interpretability. However, most existing Neuro-Symbolic Models are built on a small neural component and confined to a specific task domain, as several key challenges block scaling up.

\textbf{Challenge 1: Construction of a universal, general-purpose symbolic system. } Human experts can build symbolic components from scratch for a narrow, closed use case and domain, such as logic rules for math theorem proving. This is intractable, however, for a universal symbolic system covering as many domains as a general-purpose LLM, as shown by the Cyc project \cite{cyc}, started in the 1980s but paused in recent years of ``information explosion''. A promising alternative is to build upon the enormous existing symbolic systems from different domains and use data-driven approaches to align, merge and extend them into a colossal one matching the scope of LLMs, which we are studying concurrently.

\textbf{Challenge 2: Integration of neural and symbolic components. } Although the Type-3 Neuro-Symbolic Models above give a sketchy prototype of neural-symbolic fusion, existing instances usually simplify into a rule-based, fixed workflow template that only works on a specific task, and a universal workflow pattern for designing general Neuro-Symbolic Models seems difficult to characterize. A compromise is to abstract a shared workflow template for a task category and build Neuro-Symbolic Models for it, such as the Matching Neural Network for Information Retrieval by \citet{cheng2024}, though the shared symbolic structure of the workflow is usually oversimplified.

\textbf{Challenge 3: The boundary between neural and symbolic components for appropriate generalization. } The aforementioned difference in information compression echoes a classical philosophical topic on knowledge itself. Some knowledge is governed by a continuous structure of similarity: whether two unseen words are near-synonyms is judged by graded similarity rather than by any finite rule. \citet{Wittgenstein1953-WITPI-4} coined the term ``family resemblance'' for such knowledge, which neural networks naturally represent. Other knowledge is governed by discrete, exact rules: the correctness of a hash or of formal syntax is rule-defined, and any minor departure is simply wrong rather than approximately right, which symbolic models capture well. Characterizing knowledge precisely is therefore key to determining the proper boundary between neural and symbolic components in a Neuro-Symbolic Model.

\section{Conclusion}

We argued for a paradigm shift in AI interpretability research from explaining trained models (observation) to building auditable models (intervention). Generative interpretability formalizes this: an architecture exposes semantically meaningful, human-understandable, and causally intervenable checkpoints along its inference pass, and Neuro-Symbolic Models instantiate this paradigm. This shift is not optional in agentic AI settings, especially in high-stakes scenarios such as healthcare, legal systems and autonomous driving, where the attribution of obligations and responsibility matters beyond average performance on large samples. Ideally, audits at checkpoints on the inference path mark distinct loci of responsibility, so an undesired outcome traces to the step that bears it rather than to the system as a whole. This mission also requires efforts beyond technical work, including mechanisms of checks and balances on the power of AI audit, and thus sustained collaborations among universities, frontier AI labs and governments. 

\begin{acks}
I'm grateful to my Ph.D. advisor, Prof. ChengXiang Zhai for insightful discussions on Neuro-Symbolic Models and help with revision of the draft. 
\end{acks}

\bibliographystyle{ACM-Reference-Format}
\bibliography{references}

\appendix

\end{document}